\documentstyle[twocolumn,fullpage,fullname]{article}
\author{Philip G. Edmonds \\[0.5ex]
        Department of Computer Science, University of Toronto \\
        Toronto, Ontario, Canada {\sc M5S 1A4} \\
        {\tt pedmonds@cs.toronto.edu} }
\title{COLLABORATION ON REFERENCE TO OBJECTS \\ THAT ARE NOT MUTUALLY KNOWN}

\def\s#1{\hspace*{2em}\begin{minipage}[t]{5in}{\tt #1}\end{minipage}}

\def\ld#1#2{$\lambda$#1$\hspace{.08em}\cdot$\hspace{.12em}#2}
\def\x#1{\ld{X}{#1}}

\newcounter{conv}[part]
\def\conv #1 #2 #3 #4
{
\refstepcounter{conv}
\label{#3}
\vspace{6pt}
\par
\noindent {\bf Example \vspace*{-.5cm} \theconv} \\[6pt]
\begin{list}{\bf {\makebox[1em][l]{\hspace*{2pt}\small\theenumi}
                  \ifodd\theenumi #1:\else #2:\fi \hfill}}
{\usecounter{enumi}\settowidth{\leftmargin}{M#1 #2: }
                   \settowidth{\labelwidth}{M#1 #2:}
                   \setlength{\listparindent}{0 pt}
                   \setlength{\parindent}{0 pt}
                   \setlength{\topsep}{0 pt}}
}

\def\endconv
{
\end{list}
\vspace{2mm}
}

\def\noheadconv #1 #2
{
\begin{list}{\bf {\makebox[1em][l]{\hspace*{2pt}\small\theenumi}
                  \ifodd\theenumi #1:\else #2:\fi \hfill}}
{\usecounter{enumi}\settowidth{\leftmargin}{M#1 #2: }
                   \settowidth{\labelwidth}{M#1 #2:}
                   \setlength{\listparindent}{0 pt}
                   \setlength{\parindent}{0 pt}
                   \setlength{\topsep}{0 pt}}
}

\begin{document}
\maketitle

\begin{abstract}
\noindent In conversation, a person sometimes has to refer to an object
that is not previously known to the other participant.  We present a
plan-based model of how agents collaborate on reference of this sort.
In making a reference, an agent uses the most salient attributes of the
referent.  In understanding a reference, an agent determines his
confidence in its adequacy as a means of identifying the referent.  To
collaborate, the agents use judgment, suggestion, and elaboration moves
to refashion an inadequate referring expression.
\end{abstract}


\section{Introduction}

In conversation, a person sometimes has to refer to an object that is
not previously known to the other participant.
One particular situation in which this
arises is in giving directions.  For example:
\conv A B funny {}
\item Go straight ahead until you get to {\it a funny-looking building}.
\endconv
The hearer has to understand the reference well enough that when he
later reaches the building, he will recognize it as the intended
referent.

A reference of this sort is often achieved through a {\em
collaboration} between the conversants.  In such cases, the speaker has
the goal of having the hearer know how to identify an object.  The
speaker attempts to achieve this goal by building a description of the
object that she believes will give the hearer the ability to identify
it when it is possible to do so.  The hearer needs to be confident that the
description will be adequate as a means of identifying the referent,
but because of the inevitable differences in beliefs about the
world, he might not be.  When the hearer is not confident, the speaker
and hearer collaborate to make a new referring expression
that the hearer believes is adequate.  This can be seen in the
following portion of a telephone conversation recorded by
Psathas~\shortcite[p.~196]{psa91}.
\conv A B lowell Direction-giving
\item  Ya just stay on 2A, until ya get to Lowell Street.
\item  Is it marked?
\item  Yeah, I think there's a street sign there, it's an intersection
with lights.
\item  Okay.
\endconv
In this dialogue, speaker B is not confident that he will be able to identify
the intersection at Lowell Street, and so suggests that the intersection
might be marked.
Speaker A replies with an elaboration of the initial expression, and B finds
that he is now confident, and so accepts the reference.

This type of reference is different from the type that has been studied
traditionally by researchers who have usually assumed that the agents
have mutual knowledge of the
referent~\cite{app85b,app87,cla86,hee92,sea69}, are copresent with the
referent~\cite{hee92,coh81}, or have the referent in their focus of
attention~\cite{rei92}.  In these theories, the speaker has the
intention that the hearer either {\em know} the referent or {\em
identify} it immediately.

Although the type of reference that we wish to model does not rely on
these assumptions, we can nevertheless draw from these theories.  Thus,
we base our model on the work of Clark and
Wilkes-Gibbs~\shortcite{cla86}, and Heeman and
Hirst~\shortcite{hee92} who both modeled (the first
psychologically, and the second computationally) how people collaborate
on reference to objects for which they have mutual knowledge.  We will
briefly discuss these models, before we describe our own.

\section{Collaboration on reference}

In their fundamental experiment, Clark and Wilkes-Gibbs~\shortcite{cla86}
demonstrated that conversants use a set of inherently collaborative
procedures to establish the mutual belief that the hearer has
understood a reference.  In the experiment, two subjects were each
given a set of hard-to-describe tangram figures that were kept hidden
from the other.  One subject was required to get the other subject to
rearrange his set to match the ordering of her set, and to do so
through conversation alone.  Thus, the two subjects were obliged to
collaborate on constructing descriptions of the figures that would
allow them to be unambiguously identified; for example, {\em the one
that looks like an angel with a stick.}

Clark and Wilkes-Gibbs developed the following process model to explain their
findings.
To initiate the process, speaker A presents an initial version of a referring
expression on which speaker B passes judgment.  B can either {\em accept} it,
{\em reject} it, or {\em postpone} his decision until later.  If B
rejects or postpones, then the expression must be refashioned by either
A or B.  Refashionings are accomplished in three main ways:  {\em
repairing} the expression by correcting speech errors, {\em expanding}
the expression by adding more qualifications, or {\em replacing} part
or all of the expression with new qualifications.
Each judgment/refashioning pair
operates on the current referring expression, replacing it with a new
one.
This process continues until the expression, kept in the
participants' common ground, is mutually accepted.

This excerpt from Clark and Wilkes-Gibbs's data illustrates rejection
(line 2), replacement (line 2), and acceptance (lines 3 and 4):
\conv A B clark {}
\item Okay, and the next one is the person that looks like they're
carrying something and it's sticking out to the left.  It looks
like a hat that's upside down.
\item The guy that's pointing to the left again?
\item Yeah, pointing to the left, that's it!  {\em [laughs]}
\item Okay.
\endconv

Heeman and Hirst~\shortcite{hee92} rendered Clark and Wilkes-Gibbs's
model computationally by casting it into the planning paradigm.  Their
model covers both the {\em initiator} of a referring action, and the {\em
recipient} who tries to understand the reference.
In this model, the {\em initiator} has the goal of
having the {\em recipient} identify the referent, and so constructs a
referring plan
given a set of beliefs about what the recipient believes.
The result of the initiator's plan is a set of {\em surface speech actions},
and hearing only these actions, the recipient tries to infer a
plan in order to understand the reference.  Thus, referring expressions
are represented as plan derivations, and an unsuccessful referring
expression is an invalid plan in whose repair the agents collaborate.

An agent can infer a plan even if it is invalid in that agent's
view~\cite{pol90}.  The evaluation process attempts to find an
instantiation of the variables such that all of the constraints are
satisfied and the mental actions executable with respect to the
hearer's beliefs about the speaker's beliefs.

If the recipient finds the initial referring expression plan invalid, then
the agents will collaborate in its repair. Heeman and Hirst used plan
repair techniques to refashion an expression, and used discourse plans,
or meta-plans, to communicate the changes to it.  Thus, a collaborative
dialogue is modeled in terms of the evolution of the referring plan.

First, an agent must communicate that she has not understood a plan.
Depending on how the referring plan constrains the choice of referent,
she constructs an instance of either {\tt reject-plan} or {\tt
postpone-plan}, whose resulting surface speech actions are {\tt
s-reject} and {\tt s-postpone} respectively.
Next, one agent or the other must refashion the referring expression
plan in the context of the judgment by either replacing some of its
actions (by using {\tt replace-plan}) or by adding new actions to it
(by using {\tt expand-plan}).  The result of both plans is the surface
speech action {\tt s-actions}.

Because the model can play the role of both the initiator and the
recipient, and because it can perform both plan construction and
inference, two copies of the model can converse with one another,
acting alternately as speaker and hearer.  Acting as hearer, one copy of
the system
performs plan inference on each set of surface speech actions that it
observes, and updates the state of the collaboration.  It then switches
roles to become the speaker, and looks for a goal to adopt, and
constructs a plan that achieves it.  After responding with the surface
actions of the plan, it updates the state of the collaboration,
presupposing that the other copy will accept the plan.  The system
repeats the process until it can find no more goals to adopt, at which
time it switches back to being the hearer and waits for a response from
the other copy.

\section{Confidence and salience}

A crucial assumption of Clark and Wilkes-Gibbs's work---and of
Heeman and Hirst's model---is that the recipient of the initial
referring expression already has some knowledge of the referent in
question.  In Clark and Wilkes-Gibbs's experiments, for example, it is
one of the tangram figures.  In other words, the hearer can understand
a referring expression if its content uniquely describes an object that
he knows about.  Obviously, an agent cannot use this criterion to
understand the reference to the {\it building} in
Example~\ref{funny}---he has never heard of the building before.  What
criteria, then, does he base his understanding on?

The basis of our model is that the hearer can accept a referring
expression plan if (1) the plan contains a description that is {\em
useful} for making an {\em identification plan} that the hearer can
execute to identify the referent, and (2) the hearer is {\em confident}
that the identification plan is {\em adequate}.

The first condition, originally described by Appelt~\shortcite{app85c}, is
important because the success of the referring action depends on the hearer
formulating a useful identification plan.  We take the referring
expression plan itself to be the identification plan.  The mental
actions in the plan will encode only useful descriptions.
For the second condition to hold, the hearer must believe that the
identification plan is good enough to uniquely identify the referent
when it becomes visible.  This involves giving enough information by
using the most {\em salient} attributes of the referent.

In our model, each agent associates a numeric {\em confidence value}
with each of the attributes in the referring expression, and by
composing these, computes a level of confidence in the adequacy of the
complete referring expression plan that can be interpreted as ranging
from low confidence to high confidence.  The present composition
function is simple addition, but one could envision more complex
systems to compute confidence, such as an algebra of confidence or a
non-numeric system.  If the overall confidence value exceeds some set
value, the agent's {\em confidence threshold}, then the agent believes
the plan is adequate.  That is, if the agent is the initiator, she
believes that the other will be able to understand the reference; if
the agent is the other, he believes that he has understood the
reference.

Now, the confidence value of each attribute is equivalent to its {\em
salience} within the context of the referring expression.  Salience,
for our purposes in direction-giving, is primarily visual prominence,
but can also involve identifiability, familiarity, and functional
importance~\cite{dev76,lyn60}.  One approach is to encode the salient
properties in a static hierarchy as Davis~\shortcite{dav89}, and Reiter and
Dale~\shortcite{rei92} have done.\footnote{ These models assume that all
agents have identical beliefs, which is clearly insufficient for modeling
collaborative dialogue.}
But, ideally, salience should depend on
the context surrounding the referent.  For example, the height of a
tall building would normally be salient, but not if it were
surrounded by other tall buildings.  This computation would be quite
complex, so we have adopted a middle ground between the simple
context-independent approaches, and a full-blown contextual analysis.
The middle ground involves taking the type of object into account when
choosing attributes and landmarks that relate to it.  For example,
height and architectural style can be very salient features for
describing a building, but not for describing an intersection, for which
having a sign or traffic lights is important.  This approach still
allows us to encode salience in a hierarchy, but it is dependent on the
referent.

Table~\ref{tab:salience} shows an example of a simple salience hierarchy
that an agent might have.
The hierarchy is actually a set of partial
orderings of attributes, represented by lambda expressions, indexed by
object type.
\begin{table*}
\caption{A salience hierarchy for an agent.}\label{tab:salience}
\rule{\textwidth}{0.5pt}
\tt\begin{tabbing}
salient-attribute(3, \= intersection,traffic-lights, \=
                            \x{\ld{Y}{has(X,Y)}}). \kill
salient-attribute(4, \> building,          \>
                            \x{architectural-style(X,Style)}). \\
salient-attribute(3, \> building,          \> \x{height(X,tall)}). \\
salient-attribute(3, \> intersection,      \> \x{called(X,Name)}). \\
salient-attribute(2, \> intersection,sign, \> \x{\ld{Y}{has(X,Y)}}). \\
salient-attribute(2, \> intersection,traffic-lights, \> \x{\ld{Y}{has(X,Y)}}).
\end{tabbing}
\vspace{-1.5mm}
\rule{\textwidth}{0.5pt}
\end{table*}
In the table, the confidence value of using architectural style to
describe a building is 4.  The confidence value of a tall building is
3, and so this attribute is less salient than architectural style.  The
other rows (for describing intersections) follow
similarly.\footnote{ Given information about salience, we could
construct such a hierarchy, but we do not presume that it would be easy
to know what is salient.}

Each agent has his own beliefs about salience.
It is the difference in their beliefs that leads to the
necessity for collaboration on reference.  Ideally, the initiator should
construct referring expressions with the recipients'
(believed) beliefs about salience in mind, but we have chosen to avoid this
complexity by making the simplifying assumption that the initiator is
an expert (and thus knows best what is salient).

\section{Plans for referring}

An agent uses his salience hierarchy for two related purposes: the
first to determine what is salient in a particular situation, and the
second to determine the adequacy of a description.  So, the hierarchy
is accessed during both plan construction and plan inference.

In plan construction, the hierarchy is used for constructing initial
referring expression plans, and for elaborating on inadequate plans by
allowing an agent to choose the most salient properties of the referent
first.  The agent constructs an initial referring expression plan in
almost the same way as in Heeman and Hirst's system.  Mental actions in
the intermediate plans of a referring expression plan allow the speaker
to choose the most salient attributes that have not yet been chosen,
and constraints in the surface speech actions make sure the speaker
believes that each attribute is true.\footnote{ In Heeman and Hirst's
model, an attribute has to be mutually believed to be used.  Here,
mutual belief is not possible because the hearer has no knowledge of
the referent, but mutual belief is an intended effect of using this
plan.} Other mental actions in the intermediate plans add up the
confidence values of the attributes, and a final constraint makes sure
that the sum exceeds the agent's confidence threshold.  So, for a
referring plan to be valid, it must describe a unique object, and it
must be adequate with respect to the speaker's beliefs.  This means
that attributes beyond those required for a unique description could be
necessary.  For example, to construct the reference to the {\it
building} in Example~\ref{funny}, the speaker consulted her salience
hierarchy (in table~\ref{tab:salience}) and determined that
architectural style is salient.  Hence, she described the building as
{\it funny-looking}.  This single attribute was enough to exceed her
confidence threshold.

During plan inference, the salience hierarchy is used when evaluating a
recognized plan.  The mental actions in the intermediate plans determine
the confidence values of each attribute (from the hearer's salience
hierarchy), and add them up.  The final
constraint in the plan makes sure that the hearer's confidence
threshold is exceeded.
Thus, judging the adequacy of a referring expression plan falls out
of the regular plan evaluation process.  If the final constraint does
not hold, then the invalidity is noted so that the plan
can be operated on appropriately by the discourse plans.

For example, after recognizing the reference
in Example~\ref{funny}, the hearer evaluates the plan.  Assuming he
believes the salience information in table~\ref{tab:salience}, he
computes the confidence value of 4.  If this value exceeds his
confidence threshold, then he will accept the plan.  If not, he will
believe that there is an error at the constraint that checks his
confidence threshold.

\section{Suggestion and elaboration}

If the recipient is not confident in the adequacy of the plan, he uses an
instance of {\tt postpone-plan} to inform the initiator that he is not
confident of its adequacy, thereby causing the initiator to raise her own
confidence threshold.  Now, although he cannot refashion the
expression himself, he does have the ability to help the initiator by
{\em suggesting} a good way to expand it; {\em suggestion} is a
conversational move in which an agent suggests a new attribute that he
deems would increase his confidence in the expression's adequacy if the
expression were expanded to include the attribute.  Continuing with the
example, if the hearer were not confident about the adequacy of {\it
the funny-looking building}, he might suggest that the initiator use
height (as well as architectural style), by asking {\it Is it tall?}.
{}From this suggestion the initiator might expand her expression to {\it
the tall funny-looking building}.  So, in our sense, a suggestion is an
illocutionary act of questioning; along with actually suggesting a way
to expand a plan, the agent is asking whether or not the referent has
the suggested attribute.

To decide what suggestion to make, the agent uses an instance of {\tt
suggest-expand-plan}, which has a mental action in its decomposition
that chooses the attribute that he believes is the most salient
that has not been used already.  The result of the plan is the surface
speech action, {\tt s-suggest}, that communicates the suggestion.

However, only the initiator of the referring expression can actually
{\em elaborate} a referring expression, because only she has the
knowledge to do so.  Depending on whether the hearer of the expression
makes a suggestion or not, the initiator has two options when
elaborating a plan.  If no suggestion was made, then she can expand the
plan according to her own beliefs about the referent's attributes and
their salience.  On the other hand, if a suggestion was made, she could
instead attempt to expand the plan by affirming or denying the attribute
suggested.
If possible, she should use the suggestion to elaborate the
plan, thus avoiding unwanted conversational implicature,
but its use may not be enough to make the plan adequate.

The decomposition of {\tt expand-plan} calls the plan constructor with
the goal of constructing a {\tt modifiers} schema and with the
suggested attribute as input---in a sense, continuing the construction
of the initial referring plan.  The plan constructor attempts to find a
plan with the surface speech actions for the suggested attribute in its
yield, but this might not be possible.  In any case, the speaker
constructs an expansion that will make the plan adequate according to
her beliefs.\footnote{ Recall that she raised her confidence threshold
as a result of the hearer's postponement move, so now she must meet the
new threshold.}

The response to a suggestion depends, obviously, on whether or not the
suggestion was used to expand the plan.
The speaker can
(1) affirm that the plan was expanded with the suggestion by using the {\tt
s-affirm} speech act;
(2) affirm that the suggestion was used, along with additional
attributes that weren't suggested, by using {\tt s-affirm} and {\tt s-actions};
or
(3) deny the suggestion with {\tt s-deny}, and inform the other by {\tt
s-actions} as to how the plan was expanded.

By repeatedly using the postponement, elaboration, and suggestion
moves, the two agents collaborate through discourse on refashioning the
referring expression until they mutually believe that the recipient is
confident that it is adequate.

\section{Example}
We have implemented the model in Prolog.  Table~\ref{lowell-table}
shows the input/output of two copies of the system engaging
in a simplified version of Example~\ref{lowell}.
Note that the system generates and understands
utterances in the form of descriptions of the surface speech actions, not
surface natural language forms.  The existence of a parser and a
generator that can map between the two forms is assumed.  Complete
details of this example and of the model are given by
Edmonds~\shortcite{my-thesis}.
\begin{table*}[t]
\caption{Example of suggestion and elaboration.}
\raggedright
\rule{\textwidth}{0.5pt}
\noheadconv A B
\item Go to the Lowell Street intersection. \\
\s{s-goto(Entity) \\
   s-refer(Entity) \\
   s-attrib(Entity,\x{category(X,intersection)}) \\
   s-attrib(Entity,\x{called(X,'Lowell Street')})}
\item Does it have a sign? \\
\s{s-postpone(p1)}
\s{\begin{tabbing}
      s-suggest(p1,[\= s-attrib-rel(Entity,Entity2,\x{\ld{Y}{has(X,Y)}}), \\
                    \> s-refer(Entity2), \\
                    \> s-attrib(Entity2,\x{category(X,sign)})])
   \end{tabbing}}
\item Yes, it does, and it also has traffic lights.
\s{\begin{tabbing}
      s-affirm(p1,[\= s-attrib-rel(Entity,Entity2,\x{\ld{Y}{has(X,Y)}}), \\
                   \> s-refer(Entity2), \\
                   \> s-attrib(Entity2,\x{category(X,sign)})])
   \end{tabbing}\begin{tabbing}
      s-actions(p1,[\= s-attrib-rel(Entity,Entity3,\x{\ld{Y}{has(X,Y)}}), \\
                    \> s-refer(Entity3), \\
                    \> s-attrib(Entity3,\x{category(X,traffic-lights)})])
   \end{tabbing}}
\item Okay. \\
\s{s-accept(p123)}
\endconv
\label{lowell-table}
\rule{\textwidth}{0.5pt}
\end{table*}

\section{Conclusion}

When an agent refers to a particular object that is not previously
known to another agent, she has the intention that the agent be able to
identify the object (when it is possible to do so) by means of the
referring expression.  Because of the inevitable differences in their
beliefs about the world---specifically about what is salient---the
agents may have to collaborate to make the expression adequate.

We have implemented a computational plan-based model that accounts for
the collaborative nature of reference in the domain of interactive
direction-giving.  An agent constructs a referring expression plan by
using the referent's most salient features.  An agent understands a
reference once he is confident in the adequacy of its (inferred) plan as
a means of identifying the referent.  To collaborate, the agents use
judgment, suggestion, and elaboration moves to refashion the referring
expression until they mutually believe that the recipient has
understood.

\section*{Acknowledgments}
{\small
Our work is supported by the University of Toronto and by the Natural
Sciences and Engineering Research Council of Canada.  We are grateful to
Peter Heeman, Graeme Hirst, and Jeffrey Siskind for many helpful
discussions.
}

\bibliographystyle{fullname}
\end{document}